\pdfoutput=1
\documentclass[11pt]{article}

\usepackage[final]{acl}
\usepackage[T1]{fontenc}
\usepackage[utf8]{inputenc}
\usepackage{microtype}
\usepackage{times}
\usepackage{latexsym}
\usepackage{inconsolata}
\PassOptionsToPackage{table}{xcolor}
\usepackage{amsmath, amssymb}
\usepackage{booktabs, multirow}

\definecolor{Gainsboro}{HTML}{DCDCDC}

\usepackage{graphicx}

\newcommand{\smallwedge}{\mathop{\scalebox{0.8}{$\displaystyle\bigwedge$}}}

\title{Resource-Friendly Dynamic Enhancement Chain \\for Multi-Hop Question Answering}

\author{
Binquan Ji, Haibo Luo, Yifei Lu, Lei Hei, Jiaqi Wang,\\
{\bf Tingjing Liao, Lingyu Wang, Shichao Wang, Feiliang Ren\thanks{Corresponding Author}} \\
School of Computer Science and Engineering,\\
Northeastern University, Shenyang 110819, China \\
\texttt{jibinquan@foxmail.com} \\
\texttt{renfeiliang@cse.neu.edu.cn}
}

\begin{document}
\maketitle
\footnotetext{Code is available at \href{https://github.com/neukg/DEC}{https://github.com/neukg/DEC}}
\begin{abstract}
Knowledge-intensive multi-hop question answering (QA) tasks, which require integrating evidence from multiple sources to address complex queries, often necessitate multiple rounds of retrieval and iterative generation by large language models (LLMs). However, incorporating many documents and extended contexts poses challenges—such as hallucinations and semantic drift—for lightweight LLMs with fewer parameters. This work proposes a novel framework called DEC (Dynamic Enhancement Chain). DEC first decomposes complex questions into logically coherent subquestions to form a hallucination-free reasoning chain. It then iteratively refines these subquestions through context-aware rewriting to generate effective query formulations. For retrieval, we introduce a lightweight discriminative keyword extraction module that leverages extracted keywords to achieve targeted, precise document recall with relatively low computational overhead. Extensive experiments on three multi-hop QA datasets demonstrate that DEC performs on par with or surpasses state-of-the-art benchmarks while significantly reducing token consumption. Notably, our approach attains state-of-the-art results on models with 8B parameters, showcasing its effectiveness in various scenarios, particularly in resource-constrained environments.
\end{abstract}
\section{Introduction}
\begin{figure}[t]
  \includegraphics[width=\columnwidth]{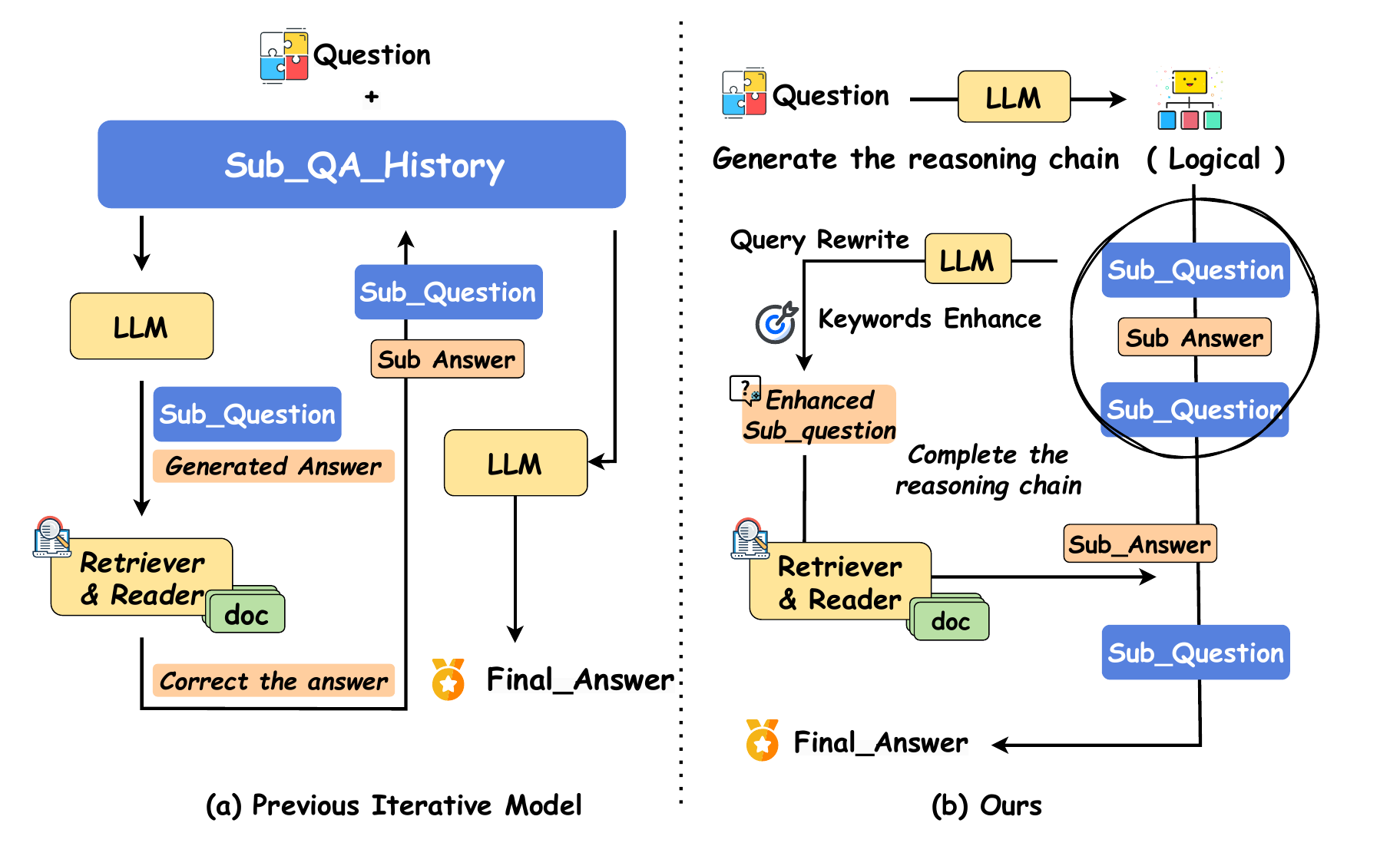}
  \caption{The process differences between our approach and the previous iterative generation framework.}
  \label{fig:introduction}
\end{figure}
In recent years, applying Retrieval-Augmented Generation (RAG) to knowledge-intensive question-answering tasks has achieved significant progress \cite{NEURIPS2020_6b493230,yu-etal-2024-chain,10.1145/3637528.3671470}. However, multi-hop question answering tasks \cite{yang-etal-2018-hotpotqa} still face a fundamental challenge: these tasks often lack a single answer document, requiring the logical decomposition of the original question into interrelated sub-questions, with the final answer derived through multi-step reasoning and cross-document retrieval \cite{10.1162/tacl_a_00667,10.1145/3589334.3645363}.
Two main approaches have emerged to address the multi-hop QA problem in RAG. The first is iterative question decomposition, which dynamically generates subquestions based on previous Q\&A interactions and retrieved documents \cite{shao-etal-2023-enhancing,trivedi-etal-2023-interleaving,press-etal-2023-measuring}. This approach is heavily dependent on the quality of the context; when lightweight language models (e.g., those with fewer than 10B parameters) process long-range contexts, semantic drift and logical discontinuities are prone to occur \cite{10.1145/3589334.3645363}. The second approach is the “generate-correct” paradigm, wherein intermediate answers are directly generated and subsequently corrected via retrieval to address hallucinations \cite{shi-etal-2024-generate,10.1145/3589334.3645363,tan-etal-2024-small}. Although these methods perform well in large-scale models such as the GPT series \cite{achiam2023gpt}, their effectiveness diminishes in lightweight LLMs. Compared to larger models, lightweight LLMs are more prone to hallucinations and exhibit greater sensitivity to erroneous information. Even minor uncorrected hallucinations during answer generation can lead to complete reasoning failures, further complicating robust reasoning.
In response to the issues inherent in the approaches mentioned above, we propose a resource-friendly Dynamic Enhancement Chain (DEC) method. As illustrated in Figure \ref{fig:introduction}, DEC first employs LLMs to decompose the original question into a sequence of semantically complete sub-questions, thereby forming a purely logical reasoning chain that avoids the injection of hallucinations at intermediate steps. Subsequently, the sub-questions are progressively refined through iterative retrieval to resolve ambiguities and supplement key reasoning information, ultimately producing accurate and reliable inferences. To enhance the efficiency of our method, we address two key challenges: \textbf{(1)} When directly decomposing the reasoning chain, the resulting sub-questions often lack effective retrieval keywords (for example, due to missing entity coreference resolution). We address this by employing a context-aware rewriting strategy that dynamically updates the sub-question formulations based on prior reasoning results to suit retrieval needs better. \textbf{(2)} To ensure retrieval precision, we integrate a lightweight module to extract discriminative keywords from the question that can differentiate key documents. During document retrieval, these discriminative keywords are used for targeted recall. This strategy improves the recall rate of gold documents at a relatively low computational cost, thereby enhancing overall model performance.
Experimental results on three multi-hop QA datasets demonstrate that DEC achieves performance comparable to or surpassing state-of-the-art benchmarks while significantly reducing token consumption. Notably, our method attains state-of-the-art results when applied to models with 8 billion parameters, underscoring its effectiveness in resource-constrained scenarios.
\begin{figure*}[t]
  \centering
  \includegraphics[width=1\textwidth]{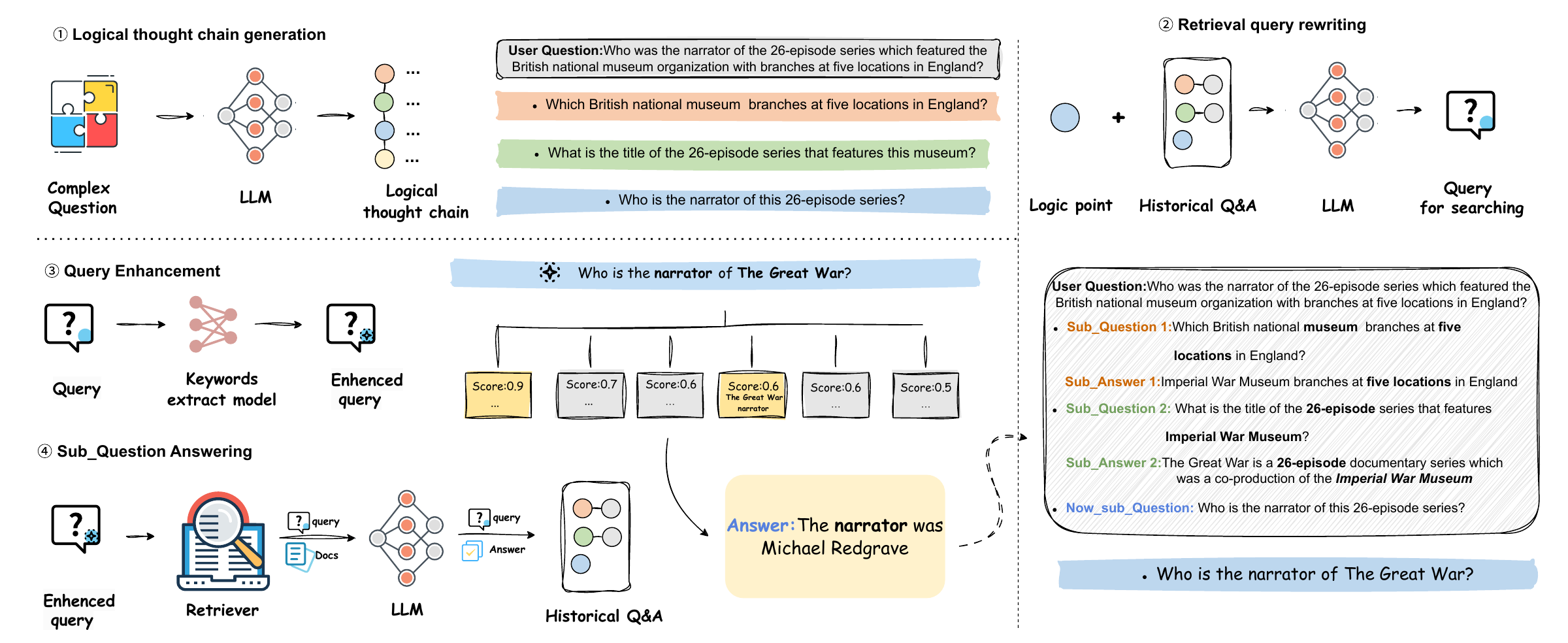}
    \caption{Workflow of the DEC framework: (a) Decompose the complex question into sub-questions expressed in natural language; (b) Reformulate the sub-questions using prior question-answer history; (c) Extract discriminative keywords for queries requiring retrieval; (d) Retrieve documents and iteratively answer all sub-questions.}
  \label{fig:framework}
\end{figure*}
\section{Related work}
\subsection{Complex Question Reasoning}
Recent advancements in LLMs have driven active research in using them to analyze, decompose, and reason through complex questions. The Chain-of-Thought (CoT) method \cite{NEURIPS2022_9d560961}, which introduces intermediate reasoning steps in prompts, marked a significant leap in LLM performance across complex tasks. \cite{NEURIPS2022_8bb0d291} further refined this approach with the "Let's Think Step by Step" prompting method, demonstrating effective multi-step reasoning in zero-shot scenarios.
Recent innovations, such as ReAct \cite{yao2023react} and Plan-and-Solve \cite{wang-etal-2023-plan}, decompose complex tasks into simpler subtasks, boosting performance in multi-step reasoning. Additionally, some CoT-based approaches integrate RAG techniques. For example, \cite{wang2024rat} proposed Retrieval-Augmented Thoughts (RAT), which refines reasoning through iterative retrieval from external knowledge sources, reducing hallucinations. The IRCoT method \cite{trivedi-etal-2023-interleaving} uses a cyclic process of retrieval and reasoning to enhance multi-hop question answering, while the Search-in-the-Chain framework \cite{10.1145/3589334.3645363} iteratively refines reasoning chains through interaction with an information retrieval system.
These advancements highlight the effectiveness of CoT methodologies in tackling complex question reasoning and lay the groundwork for the framework proposed in this study.
\subsection{Multi-hop RAG}
Multi-hop QA tasks \cite{yang-etal-2018-hotpotqa} aim to provide comprehensive answers through multi-step reasoning by integrating information from multiple sources \cite{zhang-etal-2024-end,li-du-2023-leveraging}. The use of RAG techniques \cite{NEURIPS2020_6b493230,10.1145/3637528.3671470} has become a key approach in addressing multi-hop QA questions \cite{shao-etal-2023-enhancing,asai2024selfrag,zhuang-etal-2024-efficientrag}.
A common strategy in multi-hop RAG involves iteratively generating sub-questions for decomposition \cite{trivedi-etal-2023-interleaving,press-etal-2023-measuring,shi-etal-2024-generate}. However, these methods may suffer from semantic drift due to irrelevant information, weakening the coherence of reasoning chains \cite{10.1145/3589334.3645363,shi-etal-2024-generate}. Another approach generates initial answers with potential hallucinations and corrects them through retrieval methods \cite{10.1145/3589334.3645363,tan-etal-2024-small}, but even minor uncorrected hallucinations can disrupt reasoning.
While large-scale LLMs generally exhibit stronger reasoning abilities and lower hallucination tendencies \cite{gao-etal-2023-precise,tan-etal-2024-small}, smaller models are more prone to hallucinations, complicating robust reasoning \cite{dhuliawala-etal-2024-chain,shi-etal-2024-generate}. This study introduces a framework that generates a logically coherent reasoning structure and dynamically supplements it with retrieved content, preserving logical integrity and minimizing hallucination impact.
\section{Methodology}
This section introduces the DEC method based on extended query and dynamic context augmentation, the core process of which is illustrated in Figure \ref{fig:framework}. The technique gradually resolves complex questions through a multi-stage iterative approach comprising four key steps.
Firstly, a large language model is employed to parse the user's complex question into a logically coherent chain-of-thought expressed in natural language, thereby establishing a structured framework for question decomposition.
Secondly, a dynamic query rewriting mechanism is devised to address the context dependency inherent in subsequent sub-questions. Except for the initial node, each sub-question is semantically expanded based on the cumulative QA context. In this manner, the large language model dynamically supplements any missing key information to generate an optimized query amenable to retrieval.
A precise recall method is presented to optimize retrieval performance. Before retrieval, a keyword extraction system automatically extracts distinguishing keywords from the query. During retrieval, documents are filtered based on a combination of relevance scores and keyword-matching degrees, thereby facilitating the precise recall of key evidence.
Finally, the rewritten queries and retrieved documents are submitted to the large language model to generate answers. Once an answer is obtained, the optimized query and its corresponding answer are incorporated into the QA context. This iterative process continues until every node in the initially generated logical chain-of-thought has been addressed, ultimately yielding the answer to the complex question.
\subsection{Question Decomposition and Rewriting }
To address the reasoning deviation problem in traditional iterative sub-question generation methods \cite{wang-etal-2023-query2doc} and the challenges of reasoning termination determination in small-scale language models (\ref{Experimental Results}), this paper proposes a pre-decomposed reasoning chain-based dynamic enhancement method. As shown in Figure \ref{fig:framework}, the core workflow comprises two key phases:
\begin{figure}[h]
  \includegraphics[width=\columnwidth]{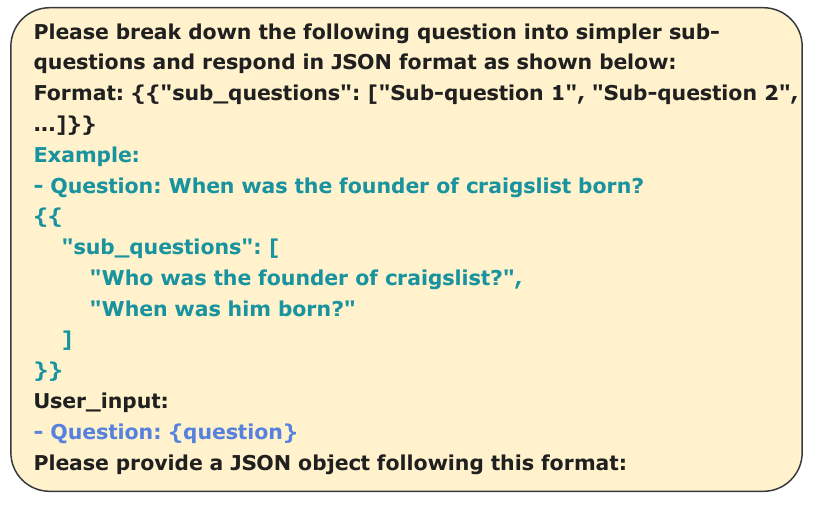}
  \caption{The instruction for the CoT generation.}
  \label{fig:QD_prompt}
\end{figure}
\noindent\textbf{Phase 1: Structured Question Decomposition} \\
Given a complex question \( Q \), a large language model \( \mathcal{M} \) parses it into a logically coherent reasoning chain \( \mathcal{C} = \{q_i\}_{i=1}^n \), where each node \( q_i \in \mathcal{C} \) represents an atomic sub-question. This process is formalized as:
\begin{equation}
   Q \stackrel{\text{CoT}}{\longrightarrow} \{q_1, \ldots, q_n\} = \mathcal{M}(Q)
\end{equation}
The CoT generation strategy is implemented via a specific prompt template (see Figure \ref{fig:QD_prompt}), ensuring the sub-question sequence \( \{q_i\} \) satisfies:
\begin{equation}
\smallwedge_{i=1}^n q_i \Rightarrow Q
\end{equation}
i.e., the conjunction of all sub-questions logically entails the original question.
\noindent\textbf{Phase 2: Dynamic context-enhanced query rewriting}
To resolve the semantic incompleteness of directly decomposed sub-questions \( \{q_i\} \) (i.e., \( q_i \) may depend on answers to preceding questions), we design a context-aware query rewriting mechanism. At the \( i \)-th iteration, the dynamic context is defined as:
\begin{equation}
\mathcal{H}_{<i} = \{(q'_j, a_j)\}_{j=1}^{i-1}
\end{equation}
where \( q'_j \) denotes the rewritten retrievable sub-question and \( a_j \) is its corresponding answer. A rewriting function \( \mathcal{R}_{rewrite} \) maps the original sub-question \( q_i \), the original question \( Q \), and historical context \( \mathcal{H}_{<i} \) into an optimized query:
\begin{equation}
q'_i = \mathcal{R}_{\text{rewrite}} \Big(\mathcal{I}_{r}, Q, q_i, \mathcal{H}_{<i} \Big)
\end{equation}
This process is guided by a designed prompt template \( \mathcal{I}_r \) (Appendix \ref{sec:Prompts}), directing the language model to:
\textbf{1)} Resolve missing referents
\textbf{2)} Inject contextual constraints
\textbf{3)} Explicitize retrieval cues
The rewritten query \( q'_i \) exhibits dual properties of \textbf{self-containedness} and \textbf{retrievability}.
\subsection{Keyword-Enhanced Precision Retrieval}
\begin{figure}[t]
  \includegraphics[width=\columnwidth]{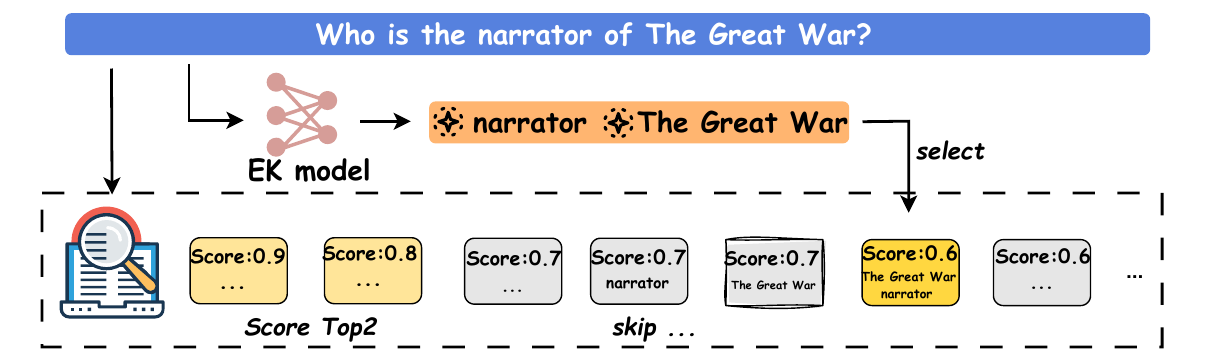}
  \caption{Demonstration of our keyword-enhanced retrieval: for a query, use the EK model to extract distinctive keywords, and then select the most relevant documents and those containing these keywords from the retrieved candidates.}
  \label{fig:EK-flow}
\end{figure}
\subsubsection{Dual-Stage Retrieval Augmentation}
To effectively complement missing information in the reasoning chain, the retriever must acquire as many relevant documents as required by the inference process. However, due to limitations of small-scale LLMs in processing long texts \cite{shi-etal-2024-generate}, we aim to maximize retrieval accuracy within constrained document quantities. We propose a keyword-enhanced query refinement method to improve retrieval precision. With the rewritten query $q'_i$ provided, the retrieval procedure is illustrated in Figure \ref{fig:EK-flow}.
\noindent\textbf{Stage 1: Discriminative Keyword Extraction}\\
A keyword extraction model $\mathcal{EK}: \mathcal{Q} \to \mathcal{K}$ is designed, where $\mathcal{K}$ denotes the keyword space. Through discriminative feature learning, this model extracts the most distinctive keyword set from query $q'_i$:
\begin{equation}
    \mathcal{K} = \{k_m\} = \mathcal{EK}(q'_i)
\end{equation}
The core design principle ensures:
\begin{equation}
\forall k \in \mathcal{K}, \quad P(k \in \mathcal{D}^* | q'_i) \gg P(k \in \mathcal{D}^{-} | q'_i)
\end{equation}
i.e., keywords exhibit significantly higher occurrence probabilities in critical documents $\mathcal{D}^*$ than in irrelevant documents $\mathcal{D}^{-}$.
\noindent\textbf{Stage 2: Hybrid Document Recall Strategy } \\
After obtaining the query keyword set \(\mathcal{K}\), we first use the retriever \(\mathcal{R}\) to perform batch retrieval for the query \(q'_i\), yielding a set of related documents \(\mathcal{D}\). The size of \(\mathcal{D}\) is relatively large, since not all documents will be used in subsequent processing. Within the document set \(\mathcal{D}\), we initially filter out documents that contain the entire keyword set \(\mathcal{K}\) and include them in the candidate document set \(\mathcal{D}^*\). To ensure that no relevant documents are missed, we also select an additional one to two documents from \(\mathcal{D}\) based on their relevance scores, supplementing the candidate document set \(\mathcal{D}^*\).
Formally, this two-stage filtering strategy is implemented as:
1. Retrieve candidate documents:
\begin{equation}
   \mathcal{D}_i = \mathcal{R}(q'_i) = \{d_j\}_{j=1}^N \quad (N=10)
\end{equation}
2. Build enhanced candidate set:
\begin{equation}
\begin{split}
       \mathcal{D}^*_i =& \underbrace{\{d \in \mathcal{D}_i | \mathcal{K}_i \subseteq \text{Terms}(d)\}}_{\text{Keyword Match}} ~~~~ \cup \\
       &\underbrace{\{d \in \mathcal{D}_i | \text{Top}_2(\mathcal{D}_i; \text{score}(q'_i, d))\}}_{\text{Relevance Backup}}
\end{split}
\end{equation}
   This guarantees $|\mathcal{D}^*_i| \geq 1$ while maintaining high relevance of retrieved results.
\subsubsection{Discriminative Keyword Model Training}
To build an efficient $\mathcal{EK}$ model, we propose a self-supervised enhanced training scheme:
\textbf{Data Construction}
Accurate keyword extraction is critical for ensuring subsequent document recall precision. Since this task is relatively simple for LLMs, we opt to train a cost-effective Llama 3.2-3B model. Specifically:
\textbf{(1)}Execute the described retrieval workflow on the HotpotQA dataset using simple prompts, relying solely on keywords for document recall.
\textbf{(2)}Validate keyword effectiveness by checking whether the retrieved documents contain the dataset-provided golden documents $d_g$.
\textbf{(3)} Define a keyword validity indicator function:
\begin{equation}
    \mathbb{I}(\mathcal{K}_i) = \begin{cases}
1, & \text{if } \exists d_g \in \mathcal{D}_{\text{gold}} \text{ s.t. } \mathcal{K}_i \subseteq \text{Terms}(d_g) \\
0, & \text{otherwise}
\end{cases}
\end{equation}
\noindent\textbf{(4)} According to the specified formula, collect effective keywords $\mathcal{K}^+ = \{k | \mathbb{I}(\mathcal{K})=1\}$ and their corresponding queries $q_t$ during iterations.
\begin{figure}[htpb]
  \includegraphics[width=\columnwidth]{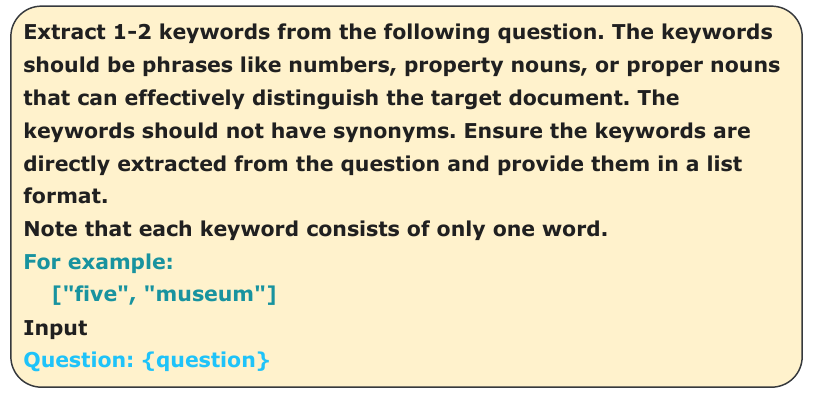}
  \caption{The instruction for the Keywords extract.}
  \label{fig:EK_prompt}
\end{figure}
\textbf{Model Fine-tuning} Based on the Llama 3.2-3B \cite{dubey2024llama} architecture, we design an instruction-tuning objective:
\begin{equation}
\mathcal{L} = -\sum_m \log P(k_m | q_t; \theta)
\end{equation}
where $\theta$ denotes the model parameters. This formulation aims to maximize the likelihood of generating discriminative keywords $\{k_m\}$ given input queries $q_t$. Consequently, it enables the model to learn question-aware keyword extraction patterns.
\section{Experiments}
This section systematically evaluates the effectiveness of the proposed method in typical multi-hop reasoning scenarios. We conducted comparative experiments on three benchmark multi-hop QA datasets and performed performance comparisons with current mainstream baseline models.
\subsection{Datasets}
This study selects three multi-hop QA benchmarks with distinct reasoning characteristics:\\
\textbf{(1) HotpotQA} \cite{yang-etal-2018-hotpotqa} requires models to perform cross-document information integration for reasoning, with question designs mandating at least two inference steps.\\
\textbf{(2) 2WikiMultiHopQA} \cite{ho-etal-2020-constructing} is constructed based on structured Wikipedia knowledge, particularly emphasizing explainable causal reasoning path modeling, providing comprehensive explanations for multi-hop questions.\\
\textbf{(3) MuSiQue} \cite{10.1162/tacl_a_00475} serves as a high-complexity benchmark, featuring question designs guaranteed through dual constraints: (a) minimum two-hop reasoning requirement; (b) answers cannot be directly obtained through single-hop retrieval.\\
Due to experimental resource constraints, we adopted a random sampling strategy to extract 500 samples from the original validation sets of each dataset to form our test set.
\subsection{Evaluation Metrics}
\label{sec:ACC}
Following the latest research paradigm in multi-hop QA (Xu et al., 2024; Shi et al., 2024), we employ a three-level evaluation framework.
\textbf{(a) Coverage Exact Match (CoverEM)}: Validates whether generated answers contain ground-truth answers through strict string matching.
\textbf{(b) Token-level F1 Score}: Calculates precision (ratio of shared tokens in predicted text) and recall (ratio of shared tokens in reference text) by counting overlapping tokens between predicted and reference texts, with F1 score computed as their harmonic mean. Token matching is based on word frequency intersection.
\textbf{(c) Semantic Accuracy (Acc†)}: To overcome limitations of rule-based metrics, we introduce Llama-3.1-8B-Instruct-based semantic evaluation \cite{shi-etal-2024-generate}. This model assesses semantic equivalence between generated and reference answers through structured prompting (see Appendix \ref{sec:Prompts} for details), effectively capturing semantic-level similarity in open-domain generation.
Additionally, in order to evaluate the resource consumption of different methods,  we record the number of sub-questions that each method generates and retrieves when addressing complex reasoning tasks, denoted as \textbf{\#SQA}. A higher \textbf{\#SQA} indicates a longer reasoning chain employed by the framework, which in turn corresponds to increased retrieval and reasoning overhead.
\begin{table*}[h]
    \centering
    \resizebox{\textwidth}{!}{
        \begin{tabular}{lcccccccccccc}
            \toprule
            \multirow{2}{*}{Method} & \multicolumn{4}{c}{\textbf{HotpotQA}} & \multicolumn{4}{c}{\textbf{2WikiMultiHopQA}} & \multicolumn{4}{c}{\textbf{MuSiQue}} \\
            \cmidrule(lr){2-5} \cmidrule(lr){6-9} \cmidrule(lr){10-13}
            & \#SQA & CoverEM & F1 & ACC† & \#SQA & CoverEM & F1 & ACC† & \#SQA & CoverEM & F1 & ACC† \\
            \midrule
            \multicolumn{13}{c}{\textbf{Llama-3.1-8B-Instruct without Retrieval}} \\
            \midrule
            Vanilla Chat & 1.00 & 23.00 & 28.81 & 24.50 & 1.00 & 23.50 & 25.71 & 19.00 & 1.00 & 4.00 & 7.08 & 4.50 \\
            Direct CoT & 1.00 & 33.00 & 35.90 & 36.00 & 1.00 & 28.00 & 30.73 & 29.00 & 1.00 & 9.00 & 14.88 & 13.00 \\
            \midrule
            \multicolumn{13}{c}{\textbf{Llama-3.1-8B-Instruct with Retrieval}} \\
            \midrule
            Direct RAG & 1.00 & 37.24 & 40.84 & 42.35 & 1.00 & 22.61 & 24.05 & 22.11 & 1.00 & 6.81 & 11.05 & 8.38 \\
            Self-Ask & 4.98 & 35.20 & 38.33 & 37.20 & 4.95 & 44.60 & 44.03 & 43.60 & 4.99 & \underline{13.23} & 16.93 & 15.23 \\
            SearChain & 3.75 & \underline{40.16} & \underline{43.06} & \underline{44.18} & 3.62 & \underline{48.60} & \underline{44.72} & \underline{43.80} & 3.73 & 13.05 & \underline{17.77} & \underline{17.47} \\
            GenGround & 5.00 & 36.20 & 39.88 & 39.60 & 5.00 & 38.40 & 35.91 & 33.60 & 5.00 & 10.40 & 15.99 & 11.60 \\
            DEC (Ours) & 3.47 & \textbf{47.19} & \textbf{50.96} & \textbf{49.60} & 3.28 & \textbf{49.18} & \textbf{46.54} & \textbf{45.70} & 3.43 & \textbf{17.21} & \textbf{20.96} & \textbf{19.26} \\
            \midrule
            \hline
            \multicolumn{13}{c}{\textbf{GPT-4o without Retrieval}} \\
            \midrule
            Vanilla Chat & 1.00 & 33.50 & 41.14 & 32.00 & 1.00 & 33.50 & 35.98 & 30.00 & 1.00 & 10.50 & 17.28 &  12.50 \\
            Direct CoT & 1.00 & 55.00 & 59.31 & 58.50 & 1.00 & 62.00 & 59.97 & 55.50 & 1.00 & \underline{25.50} & \underline{28.59} & \underline{29.50} \\
            \midrule
            \multicolumn{13}{c}{\textbf{GPT-4o with Retrieval}} \\
            \midrule
            Direct RAG & 1.00 & 57.00 & 57.39 & \underline{62.00} & 1.00 & 66.50 & 55.79 & 59.00 & 1.00 & 22.00 & 24.70 & 27.00 \\
            Self-Ask & 3.09 & 53.40 & 54.49 & 55.60 & 3.65 & 72.00 & 66.79 & \underline{67.60} & 3.44 & 20.2 & 26.17 & 22.80 \\
            SearChain & 2.24 & 53.80 & 58.12 & 58.00 & 2.90 & 63.40 & 55.66 & 54.60 & 2.62 & 23.45 & 27.77 & 27.66 \\
            GenGround & 5.00 & \textbf{60.50} & \textbf{62.37} & \textbf{63.00} & 5.00 & \underline{74.50} & \underline{68.97} & 63.00 & 5.00 & 23.00 & 24.63 & 22.50 \\
            DEC (Ours) & 3.32 & \underline{58.52} & \underline{62.11} & 60.32 & 3.41 & \textbf{78.51} & \textbf{73.26} & \textbf{71.29} & 3.22 & \textbf{27.91} & \textbf{31.30} & \textbf{30.72} \\
            \bottomrule
        \end{tabular}
    }
    \caption{Evaluation results of DEC and the baseline on three QA benchmarks. \#SQA denotes the average sub-questions generated per complex question for retrieval. ACC† indicates semantic similarity assessed by an LLM.}
    \label{tab:comparison}
\end{table*}
\subsection{Baseline Models}
The selection of baseline models is based on an incremental comparative approach, including two main categories: non-retrieval methods and retrieval-enhanced methods.\\
\textbf{Non-retrieval Baselines:} \\
\textbf{(1) Vanilla Chat}: Directly uses the original question as input to test the zero-shot question-answering capability of large language models. \\
\textbf{(2) Direct CoT} \cite{NEURIPS2022_9d560961,NEURIPS2022_8bb0d291}: Employs Chain-of-Thought techniques to guide the model in explicitly generating reasoning paths, thereby improving answer generation quality through step-by-step derivation.\\\textbf{Retrieval-Enhanced Baselines:} \\
\textbf{(3) Direct RAG}:  Constructs a single-stage retrieval-enhanced framework that performs dense retrieval on the input question, selecting the top 10 most relevant documents as contextual input for the model.  \\
\textbf{(4) Self-Ask} \cite{press-etal-2023-measuring}: Utilizes an explicit question generation mechanism to transform composite questions into a sequence of sub-questions, thereby establishing a transparent and controllable multi-step reasoning architecture.  \\
\textbf{(5) SearChain} \cite{10.1145/3589334.3645363}: Builds a dynamic retrieval-generation interaction chain that addresses the challenges of real-time knowledge updates through iterative context augmentation.\\
\textbf{(6) GenGround} \cite{shi-etal-2024-generate}: Utilizes a two-stage framework consisting of hypothesis generation and retrieval verification. In the first stage, candidate answers are generated, followed by retrieval in the second stage to correct these candidates, effectively mitigating error propagation.
\begin{table*}[t]
    \centering
    \resizebox{\textwidth}{!}{
        \begin{tabular}{lcc cc cc}
            \toprule
            \textbf{method} & \multicolumn{2}{c}{\textbf{HotpotQA}} & \multicolumn{2}{c}{\textbf{2WikiMultiHopQA}} & \multicolumn{2}{c}{\textbf{MuSiQue}} \\
            \cmidrule(lr){2-3} \cmidrule(lr){4-5} \cmidrule(lr){6-7}
            & CoverEM & F1 & CoverEM & F1 & CoverEM & F1 \\
            \midrule
            DEC (Ours)       & 47.19 & 50.96 & 49.18 & 46.54 & 17.21 & 20.96 \\
            ~~w/o EK          & 46.00(\text{↓2.5\%}) & 49.04(\text{↓3.7\%}) & 42.60(\text{↓13.3\%}) & 41.40(\text{↓11.0\%}) & 16.87(\text{↓1.9\%}) & 20.42(\text{↓2.6\%}) \\
            ~~w/o QD          & 44.44(\text{↓5.8\%}) & 45.97(\text{↓9.8\%}) & 35.68(\text{↓27.5\%}) & 38.28(\text{↓17.7\%}) & 12.24(\text{↓28.9\%}) & 17.40(\text{↓17.0\%}) \\
            ~~w/o QR          & 36.47(\text{↓22.7\%}) & 40.56(\text{↓20.4\%}) & 27.80(\text{↓43.5\%}) & 28.54(\text{↓38.7\%}) & 8.74(\text{↓49.2\%}) & 14.78(\text{↓29.5\%}) \\
            ~~w/o COT         & 39.00(\text{↓17.3\%}) & 43.18(\text{↓15.3\%}) & 26.80(\text{↓45.5\%}) & 27.82(\text{↓40.2\%}) & 8.62(\text{↓49.9\%}) & 12.95(\text{↓38.2\%}) \\
            Self-Ask        & 35.20 & 38.33 & 44.60 & 44.03 & 13.23 & 16.93 \\
            Self-Ask$_{w/EK}$   & 43.40 & 47.69 & 47.28 & 45.39 & 15.07 & 17.70 \\
            \bottomrule
        \end{tabular}
    }
    \caption{Results of the ablation study conducted on Llama-3.1-8B-Instruct. "EK", "QD", "QR", and "COT" denote keyword extraction-enhanced retrieval, structured question decomposition, dynamic query rewriting, and the combination of QD and QR modules with Chain-of-Thought reasoning, respectively. }
    \label{tab:performance}
\end{table*}
\subsection{Implementation Details}
We conduct experiments on both mainstream closed-source LLM and lightweight LLM to demonstrate the generalization performance of our approach. Specifically, we validate our method on the commercial GPT-4o model \cite{hurst2024gpt} as well as on the open-source Llama-3.1-8B-Instruct model \cite{dubey2024llama}. Regarding the construction of the knowledge base, HotpotQA and 2WikiMultiHopQA utilize the Wikipedia snapshots provided by the dataset creators, whereas MuSiQue—lacking an officially curated knowledge base—adopts the 2020 Wikipedia version from 2WikiMultiHopQA. The retrieval system employs the E5-base dense retrieval model \cite{wang2022text}.
More implementation details are provided in Appendix \ref{sec:Implementation Details}.
\subsection{Experimental Results}
\label{Experimental Results}
As shown in Table 1, the DEC framework demonstrates superior or competitive performance across models of varying parameter scales (Llama-3.1-8B-Instruct and GPT-4o) and three multi-hop reasoning benchmark datasets. Experimental results validate the significant effectiveness and strong generalization capability of the proposed multi-hop reasoning architecture. Through in-depth analysis, we derive the following key findings:
\textbf{Correlation Between Model Capacity and Hallucination Suppression}
   The closed-source GPT-4o exhibits exceptional zero-shot reasoning capabilities in the retrieval-free Direct CoT method, significantly outperforming Llama-3.1-8B (e.g., a 34\% gap in CoverEM on the 2WikiMultiHopQA dataset). Notably, GPT-4o achieves a 37.95\% reduction in reasoning chain length compared to Llama-3.1-8B (HotpotQA dataset/Self-Ask method) while improving CoverEM by 18.2\%. This phenomenon confirms the positive correlation between model parameter scale and reasoning accuracy: expanding model capacity enhances semantic understanding depth and logical coherence, thereby reducing error-prone reasoning path generation and suppressing hallucination.
\textbf{Quantitative Comparison of Reasoning Mechanism Efficiency}
   Compared to iterative reasoning baselines (Self-Ask, GenGround), the DEC framework reduces reasoning chain length by 27\% on average for Llama-3.1-8B while maintaining overall performance superiority. This discrepancy highlights two core advantages of the single-stage reasoning chain generation mechanism: (1) mitigating semantic deviation in intermediate steps through logical chain-of-thought; (2) avoiding error accumulation effects inherent in multi-step iterative generation, particularly critical for resource-constrained lightweight models.
\textbf{Synergistic Gains from Retrieval-Generation Coordination}
   On Llama-3.1-8B, the performance advantage of DEC over hypothesis-refinement methods (SearChain, GenGround) validates the effectiveness of our structured problem decomposition and dynamic query rewriting approach. By reducing the output of untrusted information, we successfully suppressed hallucination generation in lightweight models. On the 2WikiMultiHopQA dataset, DEC achieves a 12.1\% improvement in semantic accuracy (ACC†) over GenGround for Llama-3.1-8B, significantly exceeding the 8.29\% gain observed with closed-source models. This finding suggests that the proposed method provides better generalization capabilities, particularly in enhancing the performance of resource-constrained models.
\section{Further Analyses}
\subsection{Ablation Study}
To validate the effectiveness of the modules in the DEC framework, we conducted experiments by removing individual modules or key methods from the framework (see Appendix \ref{sec:Ablation Settings} for more details). The experimental results demonstrate the effectiveness of the proposed method design from the following three perspectives:
\textbf{(1) Impact of Keyword Extraction on Retrieval Quality}
When the keyword extraction module was removed (w/o EK), the performance metrics on three datasets showed a significant decrease (1.9\%-13.3\%). This indicates that extracting discriminative keywords through the EK model can effectively focus on the core retrieval needs, avoiding document noise caused by generic vocabulary.
\textbf{(2) Synergistic Effect of Question Rewriting and Reasoning Chain Decomposition}
When the structured question decomposition module is removed (w/o QD), performance drops by 5.8\% to 28.9\%, indicating that the process of breaking down complex questions into structured reasoning chains provides essential contextual dependencies and logical constraints for subsequent question rewriting and retrieval.
When the question rewriting module is removed (w/o QR), performance declines by 22.7\% to 49.2\%, confirming the critical role of the dynamic question rewriting mechanism. By incorporating the question-answer history to supplement implicit semantics, it significantly enhances both the completeness and retrieval relevance of sub-questions generated through structured decomposition.
Furthermore, removing both the question decomposition and question rewriting modules—thereby disabling the dynamic chain-of-thought construction approach (w/o COT)—results in a performance degradation of up to 49.9\%. This underscores the critical role and effectiveness of the synergistic interaction between these two components.
\textbf{(3) Performance Comparison between Structured Decomposition and Iterative Methods  }
A comparison between Self-Ask and its enhanced version Self-Ask w/EK revealed:
1. Introducing keyword extraction improved Self-Ask’s performance across all three datasets (2.5\%-15.6\%), demonstrating the effectiveness and generalizability of our retrieval strategy.
2. However, DEC still maintains a significant advantage over Self-Ask w/EK (MuSiQue F1 +18.4\%), suggesting that reasoning chain decomposition based on priors can more systematically plan the problem-solving path, avoiding path deviation and semantic accumulation errors commonly encountered in iterative methods.
\begin{figure}[htbp]
  \includegraphics[width=\columnwidth]{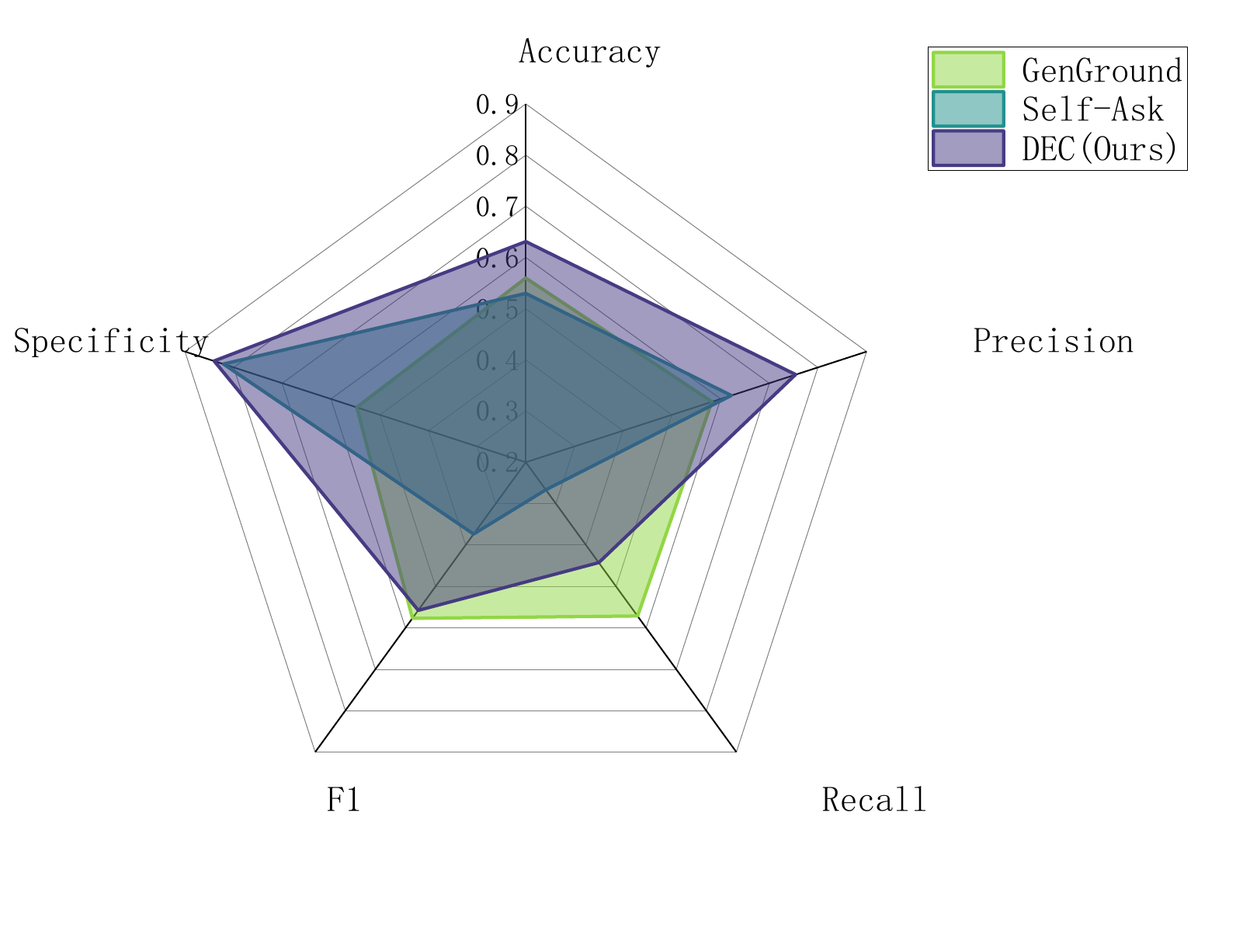}
  \caption{Performance metrics of DEC and two baselines on MuSiQue\_full for unanswerable questions.}
  \label{fig:Musique}
\end{figure}
\subsection{Performance Advantages}
\textbf{Advantages in Handling Unanswerable Queries}  In RAG tasks—where stringent reliability is paramount—accurately identifying unanswerable queries is crucial. Our experimental results on the MuSiQue\_full dataset demonstrate that the proposed DEC method significantly outperforms existing approaches in dealing with unanswerable questions. Specifically, the DEC method achieves an accuracy of 63.13\% in answerability prediction, compared to 56.00\% for GenGround and 53.00\% for Self-Ask, while also yielding a more balanced performance with a precision of 75.41\% and an F1 score of 55.76\%.
This further confirms our approach's advantages in reducing hallucinations and enhancing judgment accuracy on unanswerable questions.
\begin{figure}[h]
  \includegraphics[width=\columnwidth]{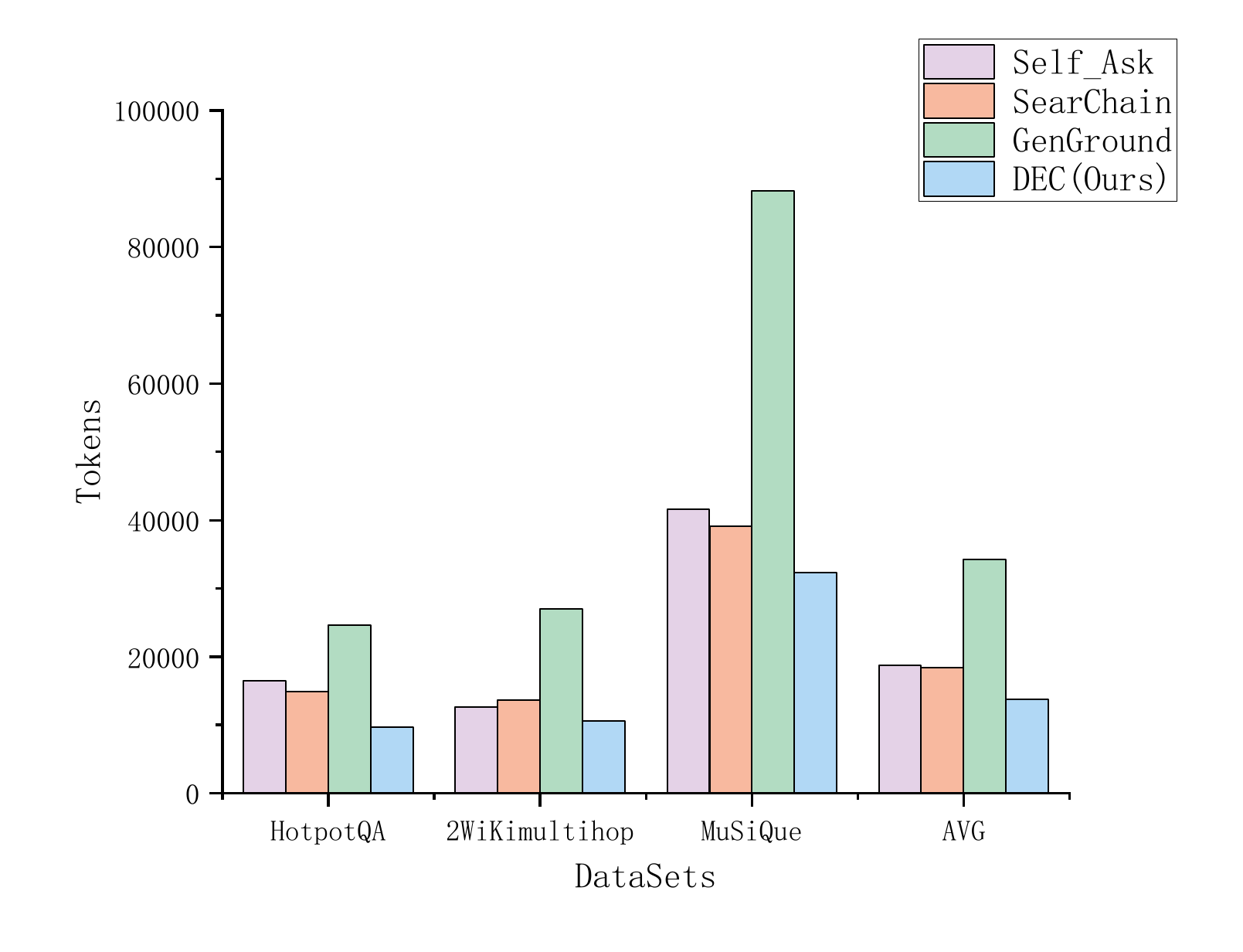}
  \caption{Statistics on the average number of tokens consumed to correctly answer a question for DEC and three baselines across three datasets.}
  \label{fig:Tokens}
\end{figure}
\vspace{-0.5cm}
\noindent\textbf{Saving Computational Resources} In parallel with our main experiments, we also evaluated the resource efficiency of different methods by recording the total token consumption for each dataset task. Specifically, we computed the total token consumption of Llama 3.1-8B-Instruct for each task and normalized it by the number of correct answers (i.e., achieving a semantic accuracy of 1). The resulting metric, \textbf{ATC}, represents the \textbf{a}verage \textbf{t}oken \textbf{c}onsumption per correct answer, allowing us to assess the trade-off between accuracy and resource expenditure. As shown in Figure \ref{fig:Tokens}, our method consistently consumes fewer tokens per correct answer across all tasks, demonstrating superior resource efficiency while maintaining answer quality.
\section{Conclusions}
In summary, our study presents a novel RAG paradigm. The DEC utilizes LLMs to directly generate logical reasoning chains, thereby minimizing the introduction of hallucinated information. It also employs iterative query rewriting to incorporate key details that may have been overlooked during the reasoning and querying processes. Furthermore, in the document retrieval phase, we implement a query strategy enhanced by discriminative keywords, effectively improving the recall rate of crucial documents. Experimental results indicate that across multiple datasets and assessments using commercial closed-source and lightweight LLMs, our method achieves—or even exceeds—the performance of existing approaches while consuming significantly fewer computational resources. Meanwhile, our approach also excels in identifying unanswerable questions. These advantages are particularly notable when applied to lightweight models with lower parameter counts.
\section*{Limitations}
Although the framework proposed in this paper demonstrates encouraging results, there are still some limitations:
Our approach heavily relies on the high-quality decomposition and knowledge supplementation of complex questions, especially when dealing with multiple sub-questions. The effectiveness of this process depends on the semantic understanding of the initial question and the accurate supplementation of relevant knowledge. If the initial decomposition or knowledge supplementation is insufficient or biased, it may lead to failure in subsequent reasoning and retrieval, thus affecting the accuracy of the final answer.
Although we have achieved significant performance improvements on lightweight LLMs, the gap in reasoning and expressive capabilities due to differences in model scale remains substantial. When faced with highly complex questions, lightweight models in our approach still struggle to reach the performance of LLMs with a higher parameter count.
\section*{Ethics Statement}
This study leverages large-scale language models to implement decomposition of reasoning chains and enhancement of retrieval, aiming to improve the response effectiveness for multi-hop questions. Throughout the research process, we strictly adhere to academic ethical standards to ensure the rigor and effectiveness of the work. With the exception of the GPT series models, all datasets, models, and methods used are publicly available and free to access, providing a high level of transparency and reproducibility for the experiments. We strive to use open-source data and frameworks to minimize potential biases as much as possible and promote fairness. Meanwhile, we ensure that our research does not harm any individual or group, nor does it involve any form of deception or information misuse.
\section*{Acknowledgments}
This project is funded by Zhiyuan Laboratory (NO. ZYL2024021) and CCF-Zhipu AL Large Model Fund (NO.202221).
\bibliography{custom}
\appendix
\section{Implementation Details}
\label{sec:Implementation Details}
To ensure experimental generalizability and reproducibility, both the generator and retriever components were implemented with default parameter configurations. Specifically, we deployed the Llama model on a single NVIDIA A6000 GPU using the vLLM \cite{kwon2023efficient} framework, while the E5 retriever was constructed through the FlashRAG \cite{FlashRAG} framework.
For the number of retrieved texts, we have made effort to adhere to the original configurations of the respective methods to ensure fairness. Specifically, Self-Ask selects the top three most relevant documents per iteration; GenGround employs a batch verification strategy by selecting three documents based on relevance per batch, with a maximum of three batches (i.e., a total of nine documents); SearChain, due to methodological constraints, selects only the single most relevant document for answer modification in each iteration; the DEC method selects the two most relevant documents from the top 10 along with documents that match specific keywords; and Direc RAG uses the top 10 most relevant documents per iteration.
For the training of the EK model, we employ the Llama-3.2-3B-Instruct architecture \cite{dubey2024llama} as the base model, utilizing Low-Rank Adaptation (LoRA) for efficient parameter fine-tuning with a learning rate of 5e-5 and training for two batches. We completed the aforementioned training using the LLaMA Factory \cite{zheng2024llamafactory} framework. The model achieves convergence after 30 minutes of training on a single NVIDIA A6000 GPU, leveraging a dataset of 1,000 multi-hop question answering samples.
\section{Ablation Study Details}
\label{sec:Ablation Settings}
To thoroughly investigate the roles of the key components in our proposed DEC method, we designed the following ablation experiments:
\textbf{w/o EK}: In this variant, the keyword extraction step is removed and document retrieval is performed using only the rewritten query (selecting the top three most relevant documents each time). This configuration is intended to verify the role of the keyword extraction module in filtering discriminative retrieval cues.
\begin{table}[b]
  \centering
  \resizebox{0.45\textwidth}{!}{
  \begin{tabular}{lcc}
    \toprule
    \textbf{Metric}                      & \textbf{DEC} & \textbf{w/o EK} \\
    \midrule
    Total Docs                      & 1218                  & 1218                      \\
    Successful Matches                & 1009                  & 861                        \\
    Document Matching Ratio           & 82.84                & 70.69 (\text{↓14.7\%})                   \\
    Fully Recalled Questions          & 328                   & 233                       \\
    Fully Recalled Ratio               & 65.60                & 46.60 (\text{↓29.0\%})                   \\
    \bottomrule
  \end{tabular}
  }
    \caption{Comparison of Retrieval Performance with and without the EK Model on the 2WikiMultihopQA Dataset}
      \label{tab:ek_performance}
\end{table}
To validate the improvement of the EK model on the recall rate of golden documents, we conducted systematic comparative experiments on the 2WikiMultihopQA dataset. As shown in Table \ref{tab:ek_performance}, after applying the EK model for keyword extraction–enhanced retrieval, the number of successfully matched documents increased significantly from 861 (70.69\%) with the baseline method to 1009 (82.84\%). In particular, in the full recall evaluation at the question level, the complete recall rate of golden documents improved from 46.6\% (233/500) to 65.6\% (328/500), representing a relative increase of 40.8\%. These results indicate that the EK model effectively increases the coverage of relevant documents during retrieval, thereby significantly enhancing the likelihood of fully recalling the gold documents for complex questions.
\textbf{w/o QD}: In this variant, we remove the question decomposition module, while still preserving the iterative nature of query rewriting from the original DEC method. Specifically, without performing complex question reasoning chain decomposition, we conduct five iterations (a number that corresponds to the maximum iteration count set by many previous iterative methods), where in each iteration the original complex question is rewritten once based on the reasoning history, while retaining the keyword extraction module. This experiment aims to demonstrate the advantage of our “first performing structured question decomposition followed by query rewriting” method over the “generate iteratively while reasoning” approach, further showcasing the importance of the synergistic effect between query rewriting and question decomposition for the overall framework performance.
\textbf{w/o QR}: Here, the question rewriting step is omitted, and retrieval relies solely on the sub-questions generated in the initial reasoning chain, while all other procedures remain identical to those in DEC. This variant examines the effectiveness of the question rewriting module in enriching query information and integrating the QA history.
\textbf{w/o COT}: In this setup, the reasoning chain generation step is excluded. Instead, document retrieval and answer generation are carried out directly using the original complex query in conjunction with keyword extraction. This setting validates the importance of decomposing the question into a multi-step reasoning chain for enhancing both retrieval and inference performance.
Furthermore, to demonstrate the superiority of our strategy—generating a reasoning chain prior to rewriting the query—we also compare against the typical iterative sub-question generation method, Self-Ask, as well as a variant of Self-Ask that incorporates a keyword extraction retrieval module (denoted as Self-Ask w/EK).
\section{Performance evaluation details}
\subsection{Handling Unanswerable Questions}
In order to conduct a comprehensive evaluation of the performance disparities between our model and baseline approaches for unanswerable question detection, we selected the MuSiQue-Full subset from the MuSiQue benchmark dataset as the experimental platform. This subset is particularly characterized by its construction of contrastive pairs that consist of both answerable and unanswerable questions. In contrast to the MuSiQue-Ans dataset, which solely comprises answerable questions, the MuSiQue-Full subset introduces unanswerable contrastive questions, thereby establishing a more stringent evaluation setting. This setup effectively mitigates the risk of model exploitation via irrelevant reasoning paths, providing a more robust foundation for assessing the model’s multi-hop reasoning capabilities and overall robustness. Due to computational resource constraints, we randomly selected 200 multi-hop questions from the validation set of MuSiQue-Full to form the test set for this experiment.
Besides the standard binary classification metrics, including Accuracy, Precision, Recall, F1-Score, and Specificity, we also introduce two supplementary metrics to assess the model's accuracy in answering questions and distinguishing between correct and incorrect responses.
\begin{table}[ht]
    \centering
    \resizebox{0.47\textwidth}{!}{
        \begin{tabular}{lccc}
        \toprule
        \textbf{Eval. Metrics} & \textbf{Self Ask} & \textbf{GenGround} & \textbf{DEC (Ours)} \\
        \midrule
        Accuracy & 53.00 & 56.00 & \textbf{63.13} \\
        Precision & 62.22 & 58.25 & \textbf{75.41} \\
        Recall & 26.67 & \textbf{57.14} & 44.23 \\
        F1-Score & 37.33 & \textbf{57.69} & 55.76 \\
        Specificity & 82.11 & 54.74 & \textbf{84.04} \\
        C Acc & 21.43 & 35.00 & \textbf{58.70} \\
        O Acc & 42.00 & 36.50 & \textbf{53.54} \\
        \bottomrule
        \end{tabular}
    }
    \caption{MuSiQue-Full Evaluation Metrics Comparison}
    \label{tab:Unanswerable}
\end{table}
\textbf{(1) Conditional Accuracy (C Acc)}
Conditional accuracy refers to the proportion of correct answers when the model predicts a question as answerable and the question is indeed answerable. It is calculated as:
\begin{equation}
    \begin{split}
        CA = \frac{\text{Correct Answers}}{TP + \text{True Answerable Subset of FP}}
    \end{split}
\end{equation}
Note: If the model predicts the question as answerable but the question is actually unanswerable (i.e., FP), then the accuracy field is invalid and must be excluded.
\textbf{(2) Overall Accuracy (O Acc)}
To comprehensively evaluate the model's end-to-end performance, we combine both the answerability prediction and the correctness of the answers. This holistic evaluation is captured by the overall accuracy metric. Overall accuracy evaluates the model's performance under the following two conditions:
\begin{itemize}
    \item Correctly abstaining from answering unanswerable questions (TN),
    \item Correctly answering answerable questions with correct answers (TP and accuracy=true).
\end{itemize}
It is computed as:
\begin{equation}
\text{Overall Accuracy} = \frac{TN + TP_{\text{acc}}}{\text{Total Number of Samples}}
\end{equation}
where \( TP_{\text{acc}} \) denotes the true positives with correct answers.
Among the aforementioned evaluation metrics, the accuracy metric for answerable questions is determined using semantic accuracy (ACC† \ref{sec:ACC}). The specific results of our method compared to the baseline in this experiment are shown in Table \ref{tab:Unanswerable}.
\subsection{Saving Computational Resources}
In the experiment, we simultaneously recorded the token consumption for each method when solving 500 questions from different datasets using Llama 3.1-8B-Instruct, as well as the average number of tokens required to correctly answer a single question (ATC). The specific data is shown in Table \ref{tab:Tokens}.
\begin{table}[ht]
    \centering
    \resizebox{0.45\textwidth}{!}{
        \begin{tabular}{lrr}
            \toprule
            \textbf{Method} & \textbf{Tok. Cons} & \textbf{ATC} \\
            \midrule
            \multicolumn{3}{l}{\textbf{HotpotQA}} \\
            Self\_Ask & 3,074,644 & 16,529.38 \\
            SearChain & 3,281,899 & 14,916.61 \\
            GenGround & 4,882,838 & 24,660.80 \\
            DEC (Ours) & \textbf{2,400,690} & \textbf{9,719.08} \\
            \midrule
            \multicolumn{3}{l}{\textbf{2WikiMultiHopQA}} \\
            Self\_Ask & 2,760,014 & 12,660.61 \\
            SearChain & 2,988,310 & 13,645.25 \\
            GenGround & 4,530,000 & 26,964.29 \\
            DEC (Ours) & \textbf{2,349,779} & \textbf{10,536.37} \\
            \midrule
            \multicolumn{3}{l}{\textbf{MuSiQue}} \\
            Self\_Ask & 3,161,428 & 41,599.00 \\
            SearChain & 3,401,388 & 39,096.14 \\
            GenGround & 5,117,786 & 88,237.69 \\
            DEC (Ours) & \textbf{3,035,764} & \textbf{32,299.21} \\
            \bottomrule
        \end{tabular}
    }
    \caption{The resource consumption of different methods.}
    \label{tab:Tokens}
\end{table}

\section{Further Analyses}
\subsection{Experiments on the 14B-Scale Model}
To further validate the generalization capability of the proposed method across different model scales, we conducted additional experiments based on the Qwen2.5-14B-Instruct model, using the same experimental setup as in the main experiments. The results are presented in Table \ref{tab:additional_comparison}. As shown in the table, the proposed method outperforms all baseline methods on nearly all evaluation metrics, fully demonstrating its effectiveness and generalization capability across different parameter scales.
\begin{table*}[h]
    \centering
    \resizebox{\textwidth}{!}{
        \begin{tabular}{lcccccccccccc}
            \toprule
            \multirow{2}{*}{Method} & \multicolumn{4}{c}{\textbf{HotpotQA}} & \multicolumn{4}{c}{\textbf{2WikiMultiHopQA}} & \multicolumn{4}{c}{\textbf{MuSiQue}} \\
            \cmidrule(lr){2-5} \cmidrule(lr){6-9} \cmidrule(lr){10-13}
            & \#SQA & CoverEM & F1 & ACC† & \#SQA & CoverEM & F1 & ACC† & \#SQA & CoverEM & F1 & ACC† \\
            \midrule
            \multicolumn{13}{c}{\textbf{Qwen2.5-14B-Instruct without Retrieval}} \\
            \midrule
            Vanilla Chat   & 1.00 & 23.50 & 29.53 & 31.00 & 1.00 & 22.00 & 25.71 & 26.50 & 1.00 &  4.00 &  7.95 &  7.50 \\
            Direct CoT     & 1.00 & 34.50 & 34.04 & 39.50 & 1.00 & 31.50 & 29.85 & 35.00 & 1.00 & 13.50 & 18.96 & \underline{24.00} \\
            \midrule
            \multicolumn{13}{c}{\textbf{Qwen2.5-14B-Instruct with Retrieval}} \\
            \midrule
            Direct RAG     & 1.00 & 51.05 & 48.33 & 58.70 & 1.00 & 50.00 & 40.75 & 44.68 & 1.00 & 12.82 & 16.96 & 20.51 \\
            Self-Ask       & 3.98 & 50.50 & 52.65 & 57.00 & 4.04 & 59.00 & 53.28 & 59.00 & 4.37 & \underline{20.00} & 22.20 & 22.50 \\
            SearChain      & 2.79 & \underline{59.00} & \underline{53.72} & \textbf{64.00} & 3.12 & 48.00 & 39.48 & 42.50 & 3.31 & 16.50 & 22.26 & 23.00 \\
            GenGround      & 5.00 & 54.00 & 57.04 & 61.50 & 4.99 & 63.00 & \underline{60.41} & \underline{63.00} & 5.00 & 17.50 & \underline{22.78} & 23.50 \\
            DEC (ours)     & 2.74 & \textbf{60.00} & \textbf{60.68} & \underline{63.00} & 2.75 & \textbf{70.35} & \textbf{66.83} & \textbf{69.85} & 3.04 & \textbf{22.56} & \textbf{25.45} & \textbf{28.72} \\
            \bottomrule
        \end{tabular}
    }
    \caption{Performance of DEC and Other Baseline Approaches on Qwen2.5-14B-Instruct}
    \label{tab:additional_comparison}
\end{table*}
\subsection{The Effectiveness of the DEC in Mitigating Model Hallucinations}
The core idea behind the DEC method is to decompose the end-to-end reasoning process into a series of relatively simple task-specific modules, generating intermediate and final outputs under explicit knowledge constraints. This design effectively reduces the risk of model hallucinations. Based on this framework, we conducted evaluation experiments along two dimensions: (1) measuring the consistency between the outputs of the keyword-extraction module and the original text; and (2) assessing the fidelity of the core content after structured question decomposition.
\subsubsection{Hallucination Assessment of the Keyword Extraction Module}
In the Discriminative Keyword Extraction (EK) module, the model’s sole responsibility is to extract retrieval-appropriate substrings from each sub-question. Because the task is so well-defined, any deviation can be directly attributed to hallucination. We quantified the hallucination rate of the EK module by calculating the match rate between the keywords extracted by the EK model and the corresponding substrings in the original sub-questions. The results are presented in Table \ref{tab:EK ACC}.
\begin{table}[htbp]
    \centering
    \resizebox{0.47\textwidth}{!}{
        \begin{tabular}{lccc}
        \toprule
        \textbf{Model} & \textbf{hotpotqa} & \textbf{2wikimqa} & \textbf{musique} \\
        \midrule
        Llama-3.1-8B & 97.39\% & 98.81\% & 96.44\% \\
        GPT-4o         & 98.31\% & 99.06\% & 97.75\% \\
        \bottomrule
        \end{tabular}
    }
    \caption{EK Model Extraction Accuracy Statistics}
    \label{tab:EK ACC}
\end{table}
As the table shows, across all datasets the EK module’s keyword-extraction match rate exceeds 96\%, indicating that the module virtually never “invents” keywords and thus has an extremely low hallucination risk.
\subsubsection{Content Fidelity after Structured Question Decomposition}
The structured decomposition module must logically reorganize the original, complex question—an evaluation that poses automated challenges. Nonetheless, the core linguistic elements before and after decomposition should remain largely unchanged, with only additional logical auxiliary words (such as interrogative words) introduced. Based on this, we removed stop words from the decomposed questions to retain the core vocabulary and then computed the proportion of these core words present in the original question. Considering factors such as variations in word forms, we allowed for some degree of fuzzy matching—deeming a word as present in the original question when its similarity exceeds 0.8. The results appear in Table \ref{tab:QD Fidelity}.
\begin{table}[h]
    \centering
    \resizebox{0.47\textwidth}{!}{
        \begin{tabular}{lccc}
        \toprule
        \textbf{Model} & \textbf{hotpotqa} & \textbf{2wikimqa} & \textbf{musique} \\
        \midrule
        Llama-3.1-8B & 84.25\% & 85.29\% & 89.21\% \\
        GPT-4o         & 92.96\% & 90.86\% & 95.33\% \\
        \bottomrule
        \end{tabular}
    }
    \caption{Problem Decomposition Fidelity Statistics}
    \label{tab:QD Fidelity}
\end{table}
It can be observed that when using the GPT-4o model, the average matching ratio of the core vocabulary in the decomposed questions compared to the original question exceeds 93\%, and even for the Llama-3.1-8B-Instruct model (where hallucinations are relatively more apparent), the average matching ratio remains above 86\%. Although hallucinations in question decomposition may have some impact, the overall performance is still within an acceptable range and sufficient for practical applications.
The two sets of experiments above confirm that DEC’s modular design combined with explicit knowledge constraints significantly reduces hallucination risk throughout the entire pipeline—from keyword extraction to question reconstruction. Even with smaller parameter models, DEC maintains high match rates and high content fidelity, ensuring the overall system’s stability and reliability.
\begin{table*}[h]
    \centering
    \resizebox{\textwidth}{!}{
        \begin{tabular}{l
                        *{3}{c}
                        *{3}{c}
                        *{3}{c}
                        *{3}{c}}
            \toprule
            \multirow{2}{*}{\textbf{Method}}
            & \multicolumn{3}{c}{\textbf{Bridge Comparison}}
            & \multicolumn{3}{c}{\textbf{Compositional}}
            & \multicolumn{3}{c}{\textbf{Comparison}}
            & \multicolumn{3}{c}{\textbf{Inference}} \\
            \cmidrule(lr){2-4} \cmidrule(lr){5-7} \cmidrule(lr){8-10} \cmidrule(lr){11-13}
            & CoverEM & F1 & ACC†
            & CoverEM & F1 & ACC†
            & CoverEM & F1 & ACC†
            & CoverEM & F1 & ACC† \\
            \midrule
            \multicolumn{13}{c}{\textbf{Llama-3.1-8B-Instruct}} \\
            \midrule
            Self‐Ask
            & 52.29 & 51.97 & 52.29
            & 29.41 & 26.83 & 29.41
            & 76.74 & 74.30 & 74.41
            & 12.06 & 22.20 &  8.62 \\
            DEC
            & 59.25 & 57.67 & 53.70
            & 42.71 & 34.03 & 37.68
            & 62.69 & 63.52 & 61.90
            & 21.81 & 31.03 & 21.81 \\
            \midrule
            \multicolumn{13}{c}{\textbf{GPT-4o}} \\
            \midrule
            Self‐Ask
            & 81.65 & 81.10 & 81.65
            & 57.35 & 46.11 & 49.50
            & 88.37 & 86.82 & 86.04
            & 68.69 & 68.05 & 63.79 \\
            DEC
            & 90.74 & 90.18 & 88.88
            & 61.27 & 48.28 & 47.05
            & 96.09 & 96.09 & 92.96
            & 77.58 & 79.20 & 75.86 \\
            \bottomrule
        \end{tabular}
    }
    \caption{Performance of DEC across Four Question Types}
    \label{tab:type_performance}
\end{table*}
\subsection{Accuracy Analysis of the DEC Method Across Question Types}
To evaluate the capability of the DEC method across different question types, we analyzed its performance on the 2WikiMultiHopQA dataset, which defines four categories of questions: bridge\_comparison, compositional, comparison, and inference. The detailed performance metrics for each category are shown in the Table \ref{tab:type_performance}.
Specifically, DEC demonstrates a notable advantage in bridge\_comparison and comparison questions, suggesting that for these types, the accurate identification and integration of relational evidence is more straightforward. In contrast, the model performs relatively worse on compositional and inference questions, indicating that these categories require more comprehensive retrieval of intermediate facts and deeper logical integration. Under such circumstances, the iterative completion process employed by DEC may still miss critical reasoning steps, resulting in lower CoverEM and F1 scores.
Nevertheless, DEC generally outperforms Self-Ask across various question types and model sizes, validating the effectiveness of our strategy for improving overall model performance.
\section{Assessing the Reliability of Smaller LLMs for Semantic Evaluation}
Given that answers in QA datasets are typically short, the associated semantic evaluation tasks are relatively straightforward. Smaller LLMs, such as Llama-3.1-8B-Instruct, are capable of producing highly reliable evaluation results. To investigate whether there is a performance gap between evaluations conducted using Llama-3.1-8B-Instruct and those using GPT-series models (e.g., GPT-4o), we conducted a comparative analysis.
Specifically, we first used Llama-3.1-8B-Instruct to evaluate the outputs of a GPT-based DEC on the HotpotQA dataset, which exhibits high accuracy, and a Llama-3.1-8B-based DEC on the MusiQue dataset, which has a higher error rate. We then used GPT-4o to re-evaluate the same model outputs under identical prompting conditions as those previously used for Llama-3.1-8B-Instruct. This ensured a fair comparison between the two evaluators under consistent evaluation settings. The results are presented in Table~\ref{tab:consistency_ratio}.
\begin{table}[h]
    \centering
    \resizebox{0.47\textwidth}{!}{
        \begin{tabular}{lc}
        \toprule
        \textbf{Data} & \textbf{Evaluation Consistency Ratio (\%)} \\
        \midrule
        DEC-GPT on HotpotQA           & 90.36\% \\
        DEC-Llama3.1-8B on MusiQue    & 95.57\% \\
        \bottomrule
        \end{tabular}
    }
    \caption{Evaluation Consistency Ratio}
    \label{tab:consistency_ratio}
\end{table}
The results demonstrate that the agreement between the two evaluation methods exceeds 90\% in both cases, supporting the validity of using Llama-3.1-8B-Instruct for semantic evaluation tasks.
\section{Prompts}
\label{sec:Prompts}
This section details the methods we employed and the prompts used during the experiments. Note that the examples given in the prompts were not part of the dataset used for testing.
\begin{table*}[ht]
\centering
\resizebox{\textwidth}{!}{
\begin{tabular}{p{0.8\textwidth}}
\toprule
\textbf{Prompt for semantic accuracy}  \\
\midrule
\footnotesize You are an experienced linguist who is responsible for evaluating the correctness of the generated responses.\\
\footnotesize You are provided with question, the generated responses and the corresponding ground truth answer.\\
\footnotesize Your task is to compare the generated responses with the ground truth responses and evaluate the correctness of the generated responses.\\
\footnotesize \#\#Example:\\
\footnotesize Example\_1:\\
\footnotesize User input:\\
\footnotesize -Question: The city where Alex Shevelev died is the capital of what region?\\
\footnotesize -Ground-truth Answer: the Lazio region\\
\footnotesize -Prediction: the answer is Lazio\\
\footnotesize Model output:\\
\footnotesize -Correctness: yes\\
\footnotesize Example\_2:\\
\footnotesize User input:\\
\footnotesize -Question: Which drink is larger, the Apple-Kneel or the Flaming volcano?\\
\footnotesize -Ground-truth Answer: The flaming volcano\\
\footnotesize -Prediction: The Apple-Kneel\\
\footnotesize Model output:\\
\footnotesize -Correctness: no\\
\footnotesize Now analyze the following question.Please be sure to output in the agreed format.\\
\footnotesize User input:\\
\footnotesize -Question: {question}\\
\footnotesize -Ground-truth Answer: \{answer\}\\
\footnotesize -Prediction: \{prediction\}\\
\footnotesize Model output:\\
\bottomrule
\end{tabular}
}
\caption{Prompt for semantic accuracy}
\label{tab:..}
\end{table*}
\begin{table*}[ht]
\centering
\resizebox{\textwidth}{!}{
\begin{tabular}{p{0.8\textwidth}}
\toprule
\textbf{Prompt for Dynamic context-enhanced query rewriting}  \\
\midrule
\footnotesize You are an auxiliary query assistant who modifies queries to better find answers to solve problems.\\
\footnotesize Follow these precise steps:\\
\footnotesize 1. \*\*Dependency Check\*\*: For each sub-question, identify if it depends on the answer to any previous sub-question. \\
\footnotesize    - State the dependency reason if it exists, otherwise, state "None".\\
\footnotesize 2. \*\*Dynamic Adjustment\*\*: Modify the sub-question to include necessary information if a dependency is present. \\
\footnotesize    - If no change is required, keep the original sub-question.\\
\small\#\#\# Input Data:\\
\footnotesize - Key\_Question:The key question that ultimately needs to be answered. The modified sub-questions should be queries that can provide crucial information for answering this question.\\
\footnotesize - Previous\_QA\_History: "The question-and-answer history of previous sub-questions, which provides crucial information for solving the key question and for the rewriting of subsequent sub-questions.\\
\small- Modifiable\_Question: The sub-questions that need to be modified.\\
\footnotesize \#\#\# Format your output as follows:\\
\footnotesize Inference\_process: Dependency reason or 'None' if not dependent\\
\footnotesize Modified\_question: Modified sub-question or original if no changes are required\\
\small\#\#Example:\\
\small- Key\_Question:When was the founder of craigslist born?\\
\small- Previous\_QA\_History:\\
\footnotesize sub\_question\_1:Who was the founder of craigslist?, sub\_answer:Craigslist was founded by Craig Newmark.\\
\small- Modifiable\_Question:"When was him born?"\\
\footnotesize Inference\_process: The sub-question "When was him born?" depends on the answer to sub-question\_1 because "him" refers to the previously identified founder, Craig Newmark.
\footnotesize Modified\_question: When was Craig Newmark born?\\
\footnotesize Now analyze the following question. Please be sure to output in the agreed format.\\
\footnotesize User input:\\
\footnotesize - Key\_Question:\{question\}\\
\footnotesize - Previous\_QA\_History:\{history\}\\
\footnotesize - Modifiable\_Question:\{sub\_question\}\\
\footnotesize Model output:\\
\bottomrule
\end{tabular}
}
\caption{Prompt for Dynamic context-enhanced query rewriting}
\label{tab:..}
\end{table*}
\begin{table*}[ht]
\centering
\resizebox{\textwidth}{!}{
\begin{tabular}{p{0.8\textwidth}}
\toprule
\textbf{Prompt for answering sub\_question}  \\
\midrule
\footnotesize Answer the following question briefly based on relevant information:\\
\footnotesize Question: \{sub\_question\}\\
\footnotesize Context: \{rel\_text\}\\
\midrule
\textbf{Prompt for reasoning through the chain of thought to the answer}  \\
\midrule
\footnotesize Synthesize an answer to the original question based on the answers to sub-questions:\\
\footnotesize "Your reasoning process should be separated into two fields from the answer. In the answer field, please provide the answer as concisely as possible. The answer should be given in the form of words or phrases as much as possible. \\
\footnotesize \#\#\# Input Data:\\
\footnotesize - Original\_Question:The key question that ultimately needs to be answered.\\
\footnotesize - Evidence:Question-and-answer pairs of the sub-questions split from the original question, which are used to answer the final original question.\\
\footnotesize \#\#\# Format your output as follows:\\
\footnotesize Inference\_process: Your reasoning process\\
\footnotesize Answer: Modified Provide answers as concisely as possible\\
\footnotesize \#\#Output Example:\\
\footnotesize Inference\_process: Based on the sub-questions and answers, I identified the series that matches the description as Animorphs, a science fantasy young adult series told in first person. The series has companion books that narrate the stories of enslaved worlds and alien species, which aligns with the nature of the companion books in the Square Enix series. \\
\footnotesize Answer: Animorphs\\
\footnotesize Now analyze the following question. Please be sure to output in the agreed format.\\
\footnotesize User input:\\
\footnotesize - Original\_Question:\{question\}\\
\footnotesize - Evidence:\{history\}\\
\footnotesize Model output:\\
\bottomrule
\end{tabular}
}
\caption{Prompt for reasoning through the chain of thought to the answer}
\label{tab:..}
\end{table*}
\begin{table*}[ht]
\centering
\resizebox{\textwidth}{!}{
\begin{tabular}{p{0.8\textwidth}}
\toprule
\textbf{Prompt for Direct CoT}  \\
\midrule
\footnotesize You are a question-answering system capable of constructing a reasoning chain based on your world knowledge. Given the input question, follow these steps to infer the answer:\par
\footnotesize 1. Break down the question and identify key facts that will help in the reasoning process.\par
\footnotesize 2. Use your world knowledge to find relevant information that can help answer the question.\par
\footnotesize 3. Build a chain of inferences leading to the final answer.\par
\footnotesize 4. Format the output as follows:\par
\footnotesize    - First, list the reasoning steps, clearly numbered.\par
\footnotesize    - Then, conclude with the final answer in the format: \par
\footnotesize      'So the final answer is: <answer>'\par
\footnotesize Example:\par
\footnotesize Question: What government position was held by the woman who portrayed Corliss Archer in the film Kiss and Tell?\par
\footnotesize Inference\_process:\par
\footnotesize 1. Kiss and Tell is a 1945 American comedy film starring then 17-year-old Shirley Temple as Corliss Archer.\par
\footnotesize 2. Shirley Temple Black was named United States ambassador to Ghana and to Czechoslovakia and also served as Chief of Protocol of the United States.\par
\footnotesize So the final answer is: Chief of Protocol\par
\footnotesize Now, given the following question, please provide the inference process and the final answer.\par
\footnotesize Question: {question} \\
\midrule
\textbf{Prompt for Direct RAG}  \\
\midrule
\footnotesize You are a question-answering system capable of combining world knowledge and information from provided documents to answer a question. Given the input question and a list of relevant documents retrieved based on the question, please follow these steps:\par
\footnotesize 1. Read through the provided documents and identify relevant information.\par
\footnotesize 2. Filter out irrelevant or redundant information and focus on the most useful content.\par
\footnotesize 3. Combine the knowledge from the documents with your own world knowledge to construct a reasoning chain.\par
\footnotesize 4. Format the output as follows:\par
\footnotesize    - First, list the reasoning steps, clearly numbered, describing how you combined the information from the documents with your world knowledge.\par
\footnotesize    - Then, conclude with the final answer in the format: \par
\footnotesize      'So the final answer is: <answer>'\par
\footnotesize Example:\par
\footnotesize Question: What government position was held by the woman who portrayed Corliss Archer in the film Kiss and Tell?\par
\footnotesize Documents:\par
\footnotesize 1. "Kiss and Tell" is a 1945 American comedy film starring Shirley Temple as Corliss Archer.\par
\footnotesize 2. Shirley Temple Black served as U.S. ambassador to Ghana and Czechoslovakia and was also appointed Chief of Protocol.\par
\footnotesize 3. The film was a major success in the 1940s, helping Shirley Temple become one of the most famous child stars of the era.\par
\footnotesize Inference\_process:\par
\footnotesize 1. The film "Kiss and Tell" featured Shirley Temple as Corliss Archer.\par
\footnotesize 2. Relevant documents indicate that Shirley Temple Black later became a U.S. ambassador to two countries and served as Chief of Protocol.\par
\footnotesize 3. Combining this with world knowledge about Shirley Temple's later career, the most relevant position she held was Chief of Protocol.\par
\footnotesize So the final answer is: Chief of Protocol\par
\footnotesize Now, given the following question and documents, please provide the inference process and the final answer.\par
\footnotesize Question: \{question\}\par
\footnotesize Documents:\{documents\}\\
\bottomrule
\end{tabular}
}
\caption{Prompt for reasoning through the chain of thought to the answer}
\label{tab:..}
\end{table*}
\end{document}